\documentclass[runningheads]{llncs}
\usepackage[T1]{fontenc}
\usepackage{graphicx}
\usepackage{multicol}
\usepackage{multirow}
\usepackage{comment}
\usepackage{booktabs}
\begin{document}
\title{Graph Neural Networks for Misinformation Detection:
Performance–Efficiency Trade-offs}
\titlerunning{GNNs for Misinformation Detection:
Performance–Efficiency Trade-offs}
%

%

\makeatletter
\newcommand{\printfnsymbol}[1]{\textsuperscript{\@fnsymbol{#1}}}
\makeatother

\author{
Soveatin Kuntur\inst{1,3}\orcidID{0000-0001-6029-941X}%
\thanks{These authors contributed equally to this work.} \and
Maciej Krzywda\inst{4}\orcidID{2222-3333-4444-5555}%
\printfnsymbol{1} \and
Anna Wróblewska\inst{1}\orcidID{0000-0002-3407-7570} \and
Marcin Paprzycki\inst{2}\orcidID{0000-0002-8069-2152} \and
Maria Ganzha\inst{1}\orcidID{0000-0001-7714-4844} \and
Szymon Łukasik\inst{4,2,6}\orcidID{0000-0001-6716-610X} \and
Amir H. Gandomi\inst{7,8,9}\orcidID{0000-0002-2798-0104}
}

\authorrunning{S. Kuntur \& M. Krzywda et al.}

\institute{
Warsaw University of Technology, plac Politechniki 1, 00-661 Warsaw, Poland \and
Systems Research Institute, Polish Academy of Sciences, Newelska 6, 01-447 Warsaw, Poland \and
Corresponding author \and
AGH University of Krakow, Faculty of Physics and Applied Computer Science, Krakow, Poland \and
NASK National Research Institute, ul. Kolska 12, 01-045 Warsaw, Poland\and
Faculty of Engineering and IT, University of Technology Sydney, 5 Broadway, Ultimo NSW 2007, Australia \and
University Research and Innovation Center (EKIK), Óbuda University, Bécsi út 96/B, Budapest, 1034, Hungary \and
Department of Computer Science, Khazar University, Baku, Azerbaijan}

\maketitle              
\begin{abstract}
The rapid spread of online misinformation has led to increasingly complex detection models, including large language models and hybrid architectures. However, their computational cost and deployment limitations raise concerns about practical applicability. In this work, \textbf{we benchmark graph neural networks (GNNs) against non-graph-based machine learning methods} under controlled and comparable conditions. We evaluate lightweight GNN architectures (GCN, GraphSAGE, GAT, ChebNet) against Logistic Regression, Support Vector Machines, and Multilayer Perceptrons across seven public datasets in English, Indonesian, and Polish. All models use identical TF–IDF features to isolate the impact of relational structure. Performance is measured using F1 score, with inference time reported to assess efficiency. GNNs consistently outperform non-graph baselines across all datasets. For example, GraphSAGE achieves 96.8\% F1 on Kaggle and 91.9\% on WELFake, compared to 73.2\% and 66.8\% for MLP, respectively. On COVID-19, GraphSAGE reaches 90.5\% F1 vs. 74.9\%, while ChebNet attains 79.1\% vs. 66.4\% on FakeNewsNet. These gains are achieved with comparable or lower inference times. Overall, the results show that classic GNNs remain effective and efficient, challenging the need for increasingly complex architectures in misinformation detection.
\end{abstract}

\keywords{Misinformation Detection \and Graph Neural Networks \and Performance--Efficiency Trade-offs \and Low-Resource Learning \and Multilingual Text Classification}

\section{Introduction}
The rapid spread of misinformation online continues to undermine public trust and democratic processes~\cite{mehta-etal-2022-tackling} \cite{rode-hasinger-etal-2022-true}. 
Recent advances in large language models (LLMs) have led to strong performance in misinformation detection. 
However, these models remain difficult to deploy in practice due to their heavy computational requirements, 
sensitivity to distribution shift, and limited multilingual robustness~\cite{kuntur2025under}. 
As documented in recent surveys, LLM-based approaches also face challenges related to hallucination, 
explainability, and reliance on extensive fine-tuning, making them unsuitable for many real-time or 
resource-constrained environments~\cite{kuntur2025under}. 
Consequently, recent research has increasingly explored complex hybrid architectures that combine 
Transformers, graph neural networks, and multi-feature fusion mechanisms to maximize detection performance, 
often at the cost of increased model complexity and computational overhead~\cite{cui2025cmgn}.

In contrast, Graph Neural Networks (GNNs) offer an appealing alternative for misinformation detection by 
explicitly modeling relational structure—such as similarity between articles, shared sources, or co-occurring 
claims -- while remaining comparatively lightweight and interpretable~\cite{scarselli2008graph}. 
However, despite substantial progress in designing increasingly sophisticated GNN variants, it remains unclear 
whether \emph{classic, computationally efficient GNNs} provide meaningful advantages over simpler non-graph 
models when evaluated under controlled conditions. Recent evidence from computational social science suggests 
that carefully designed but simple graph-based representations can remain competitive, and in some cases even 
outperform more complex embeddings, also significantly reducing computational cost~\cite{xu2026domain}. 
These findings motivate a closer examination of when relational inductive bias contributes measurable benefits 
in misinformation detection.

This gap motivates a systematic re-evaluation of classic GNN architectures, defined here as early, widely adopted, and computationally lightweight models, for misinformation detection, without any fine-tuning.
Misinformation datasets differ widely—from short political claims (LIAR) to long-form news articles 
(FakeNewsNet), from balanced corpora (WELFake) to highly imbalanced ones (COVID-19), and across multiple 
languages including English, Indonesian, and Polish. In such heterogeneous settings, it is not obvious 
when lightweight GNNs should succeed or fail, how they compare against strong non-graph baselines such as 
Logistic Regression and Multilayer Perceptrons (MLPs), or whether they offer practical performance--efficiency 
trade-offs relative to Transformer-based models.

To address these questions, we conduct a large-scale multilingual benchmarking study of classic GNNs 
(Graph Convolutional Network (GCN)~\cite{gcnkipf2016semi}, Graph Attention Network (GAT)~\cite{velickovic2018graph}, Chebyshev Network (ChebNet)~\cite{chebnet}, 
Simplifying graph convolutional networks (SGC)~\cite{sgcpmlr-v97-wu19e}, and Feature-Steered Graph Convolution (FeaStConv)~\cite{Verma_2018_CVPR}) for misinformation detection. 
We evaluate seven datasets spanning political fact-checking, clickbait detection, and pandemic-related 
misinformation, and systematically vary dataset size (10\% vs.\ full), graph sparsity (2 vs.\ 5 neighbors), 
and the use of light pre-training. Rather than proposing new architectures, our goal is to clarify the 
performance--efficiency trade-offs of classic GNNs relative to non-graph and Transformer-based models. 
Our results show that classic GNNs achieve modest but consistent improvements over Logistic Regression and 
MLP baselines under identical feature representations. Based on these considerations, we formulate the following research questions to guide our study:

\begin{itemize}
    \item \textbf{RQ1:} Are classic GNNs competitive with strong non-graph baselines for misinformation detection when evaluated under controlled and comparable settings?
    \item \textbf{RQ2:} Which classic GNN architectures perform best across heterogeneous misinformation datasets?
    \item \textbf{RQ3:} Do classic GNNs remain effective in low-resource settings with limited labeled data?
    \item \textbf{RQ4:} What are the performance-efficiency trade-offs of classic GNNs compared to non-graph models?
\end{itemize}

Rather than proposing new architectures, this work aims to clarify the practical role of classic GNNs by systematically answering these questions across multiple datasets, languages, and experimental conditions. For the purpose of anonymous review, an anonymized implementation and
supplementary materials are available at:
\url{https://github.com/mkrzywda/gnn-misinformation-tradeoffs}.

\section{Related Work}
\label{sec:related-work}

The increasing prevalence of misinformation on social media has motivated extensive research on the detection of automated fake news and disinformation. Broadly, existing approaches can be categorized into two main groups: graph-based methods, which exploit relational and structural information, and text-based models, which focus on understanding semantic content.

\paragraph{GNN-based Disinformation Detection.}
Graph Neural Networks (GNNs) have been widely adopted for misinformation detection due to their ability to model complex relationships among news articles, sources, users, and propagation patterns. A recent comprehensive survey by Lakzaei et al.~\cite{lakzaei2024disinformation} reviews GNN-based disinformation detection methods, categorizing prior work by graph construction strategies, feature representations, and architectural design. The survey highlights that most existing approaches emphasize increasingly complex architectures, heterogeneous graphs, or multimodal feature fusion, often relying on explicit social or propagation networks. While these methods achieve strong performance, the survey also identifies open challenges related to scalability, early-stage detection, and the lack of controlled evaluations that isolate the contribution of graph structure itself.

Several studies focus on architectural innovation within the GNN paradigm. For example, Phan et al.~\cite{phan2023fake} categorize GNN-based approaches into convolutional, attention-based, and autoencoder-based models, emphasizing the role of attention mechanisms and multimodal integration in improving detection accuracy. Chang et al.~\cite{chang2024graph} introduce the Graph Global Attention Network with Memory (GANM), which incorporates both local and global structural information to enhance representation learning. Models such as GraphSAGE further extend GNN capabilities through inductive learning, enabling scalability to large and evolving datasets~\cite{hamilton2017inductive}. Despite these advances, comparatively less attention has been paid to \textbf{reassessing classic, computationally efficient GNN architectures under unified and controlled experimental settings.}

\paragraph{Lightweight Graph Representations and Efficiency.}
Beyond misinformation detection, recent work in computational social science has demonstrated that carefully designed but simple graph representations can remain highly effective. Xu and Sasahara~\cite{xu2026domain} propose a domain-based user embedding method that constructs graphs from URL co-occurrence patterns, demonstrating that it outperforms text-based and retweet-based embeddings while significantly reducing computational cost across multiple datasets. Their findings suggest that relational inductive bias, even when instantiated through lightweight graph constructions, can provide meaningful benefits without relying on complex architectures. This line of work supports the premise that simplicity and efficiency should be considered first-class objectives, particularly for large-scale or real-time systems. It motivates \textbf{a systematic re-evaluation of classic GNN models.}

\section{Our Approach}\label{sec:methodology}

\subsection{Datasets}\label{subsec:datasets}

We evaluate our models across seven publicly available datasets for misinformation and clickbait detection in English, Indonesian, and Polish. The datasets vary in size, label granularity, class balance, and curation method. The curation method indicates whether labels were assigned by human annotators or generated automatically using source-based or heuristic procedures, allowing us to assess model robustness across heterogeneous settings.

LIAR provides fine-grained truthfulness annotations for short political claims and is formulated as a six-class classification task (ranging from \emph{pants-fire} to \emph{true}). It makes LIAR the only multi-class dataset in our benchmark suite, introducing an additional level of difficulty compared to binary misinformation detection. MIPD contains human-annotated Polish news articles labeled for disinformation, representing a multilingual and manually curated benchmark. CLICK-ID focuses on clickbait detection in Indonesian news headlines, capturing stylistic forms of deception rather than factual correctness.

ISOT, WELFake, and FakeNewsNet support binary fake news classification in English and differ primarily in scale and construction. While ISOT is assembled from curated sources, WELFake and FakeNewsNet are automatically aggregated from multiple publishers and repositories. The COVID-19 Fake News dataset targets domain-specific misinformation related to the pandemic, reflecting a more topical and temporally constrained distribution of information.

As summarized in Table~\ref{tab:dataset_comparison}, the benchmark suite includes both binary and multi-class classification settings, balanced and imbalanced label distributions, and a mix of human-annotated and automatically curated datasets.

\begin{table}[t]
\centering
\begin{tabular}{p{3cm} c r c p{3.5cm} c}
\toprule
\textbf{Dataset} & \textbf{Lang.} & \textbf{Size} & \textbf{Balance} & \textbf{Labels} & \textbf{Curation} \\ 
\midrule
CLICK-ID~\cite{william2020click}      & ID & 15K   & I & Clickbait/Non-clickbait & H \\
LIAR~\cite{wang-2017-liar}          & EN & 12.8K & I & Fine-grained truth       & H \\
FakeNewsNet~\cite{shu2020fakenewsnet}   & EN & 23.9K & I & Fake/Real                & A \\
Kaggle ISOT~\cite{ahmed2017detection}         & EN & 44.9K & B & Fake/Real                & A \\
WELFake~\cite{welfake9395133}       & EN & 72K   & B & Fake/Real                & A \\
COVID-19~\cite{shuja2021covid}   & EN & 10K   & I & Fake/Real                & A \\
MIPD~\cite{modzelewski-etal-2024-mipd}         & PL & 13K   & I & Disinformation           & H \\
\bottomrule
\\
\end{tabular}
\caption{\label{tab:dataset_comparison} Summary of datasets used. \textbf{Lang.}: language. \textbf{Balance}: \textbf{B} = balanced, \textbf{I} = imbalanced. \textbf{Curation}: \textbf{H} = human-annotated, \textbf{A} = automatic.}
\vspace{-1em}
\end{table}

\subsection{Data Preprocessing}\label{sec:data-prep}
Data preprocessing followed a unified pipeline to ensure consistency and accuracy. Data is loaded from CSV/TSV files, relevant columns are extracted, and categorical labels are mapped to numerical values. Stratified sampling splits the dataset into training (80\%), validation (10\%), and test (10\%) sets. Text is transformed using TF-IDF (max 5,000 features). For GNNs, a k-NN graph is constructed, where nodes represent text samples, edges encode similarity, and features are derived from TF-IDF vectors. The dataset is formatted as \texttt{torch\_geometric.data.Data}, ensuring scalability across sources.

\subsection{Non-Graph Baselines}
\label{subsec:baselines}

To isolate the contribution of graph structure from feature learning, we include three strong non-graph baselines that operate directly on the same TF--IDF representations used by the GNN models: Logistic Regression (LR), linear Support Vector Machines (SVM), and a Multilayer Perceptron (MLP).

The LR baseline is trained with L2 regularization and a maximum of 5{,}000 iterations. For SVM, we use a linear kernel and apply probability calibration to enable consistent evaluation across metrics. The MLP consists of two hidden layers with 256 and 128 units, ReLU activation, early stopping, and a maximum of 200 training iterations. All non-graph baselines are trained using the same stratified data splits as the GNN models, ensuring a controlled and fair comparison.

\subsection{Our Approach}\label{sub:experimental}

Our approach employs graph neural networks (GNNs) to detect disinformation by converting textual data into high-dimensional node features via TF-IDF vectorization and constructing graphs using k-nearest neighbor (k-NN) techniques to capture semantic relationships. We evaluate a suite of state-of-the-art GNN models, including GCN, GraphSAGE, GAT, SGC, and GIN, using a two-phase training pipeline: a brief pre-training phase followed by full training with stratified 5-fold cross-validation and early stopping.

Each text sample is transformed into a 5,000-dimensional TF-IDF vector and converted into a k-NN graph using \texttt{knn\_graph} with $K=5$ or $K=2$, controlling graph sparsity by limiting node connections to the top-2 or top-5 nearest neighbors. During pre-training (5 epochs), models minimize MSE loss on randomly assigned edge labels (0 or 1) using a dedicated \texttt{pretrain\_fully\_connected} layer, enhancing the quality of intermediate representations. The main training phase lasts up to 500 epochs, utilizing the Adam optimizer (learning rate = 0.001) and cross-entropy loss, with early stopping (patience = 10) based on the validation loss.

All experiments were conducted on a workstation equipped with an AMD Ryzen 9 7950X CPU (16 cores, 32 threads), 124~GB RAM, and two NVIDIA RTX 4090 GPUs. To evaluate generalizability and robustness, we tested each model on both the full dataset and a stratified 10\% subset, ensuring that the class distribution was preserved in all splits. Models were trained with and without the pre-training phase, and results were averaged over three runs with different random seeds. Our experiments were implemented as a \texttt{Python script} using \texttt{Python 3.12}.

We evaluate all models using standard supervised classification metrics, including Accuracy, Precision, Recall, F1 Score, Area Under the Receiver Operating Characteristic Curve (AUC-ROC), and Matthew Correlation Coefficient (MCC). Due to class imbalance in several datasets, F1 score is used as the primary metric in the main analysis, while additional metrics are reported in Appendix for completeness. 
\section{Result}\label{sec:result}
\subsection{RQ1: Are classic GNNs competitive with strong non-graph baselines?} \label{sec:RQ1}

Table~\ref{tab:main_results} compares classic GNN models with strong non-graph baselines, including Logistic Regression, SVM, and MLP, across all seven datasets. For clarity of presentation, we report for each dataset the best-performing GNN configuration, selected based on mean F1 score, alongside representative non-graph baselines. Performance is evaluated primarily using F1 score, with additional metrics reported for completeness, together with inference time to assess practical efficiency. Complete results for all evaluated GNN architectures and hyperparameter configurations are provided in the appendix.

Across all datasets, classic GNNs consistently achieve the highest F1 scores. The performance improvements over non-graph baselines are systematic rather than dataset-specific. Compared to MLPs, which already constitute a strong nonlinear baseline, classic GNNs improve F1 scores by approximately 12 to 30 percentage points, depending on the dataset, with even larger margins observed over linear models such as Logistic Regression.

Importantly, these gains in predictive performance are not accompanied by prohibitive computational costs. Inference times for GNNs remain within the same order of magnitude as those of MLPs and are substantially lower than those of SVMs on larger datasets such as Kaggle and WELFake. Overall, the results indicate that classic GNNs are not only competitive with but consistently outperform strong non-graph baselines across heterogeneous datasets, while maintaining practical inference efficiency.

\begin{table*}[t]
\scriptsize
\centering
\caption{Performance comparison across datasets. Results are reported as mean $\pm$ standard deviation over multiple runs. F1 score is the primary metric. MCC is reported to account for class imbalance. LogReg denotes Logistic Regression, MLP denotes Multilayer Perceptron, and SVM denotes Support Vector Machine.}
\label{tab:main_results}
\resizebox{\textwidth}{!}{%
\begin{tabular}{llcccccc}
\toprule
\textbf{Dataset} & \textbf{Model} &
\textbf{Precision (\%)} &
\textbf{Recall (\%)} &
\textbf{F1 (\%)} &
\textbf{AUC-ROC (\%)} &
\textbf{MCC (\%)} &
\textbf{Inference (ms)} \\
\midrule
\midrule

\multirow{4}{*}{CLICK-ID}
 & LogReg & 50.3 $\pm$ 21.3 & 50.3 $\pm$ 21.4 & 50.3 $\pm$ 21.3 & 51.5 $\pm$ 21.6 & 13.1 $\pm$ 5.3 & 5.85 \\
 & SVM    & 50.2 $\pm$ 21.4 & 51.3 $\pm$ 22.1 & 49.4 $\pm$ 21.1 & 50.9 $\pm$ 21.8 & 11.9 $\pm$ 4.6 & 6.95 \\
 & MLP    & 58.8 $\pm$ 25.6 & 58.5 $\pm$ 25.6 & 57.0 $\pm$ 24.9 & 61.9 $\pm$ 26.9 & 28.5 $\pm$ 12.3 & 10.84 \\
 & GAT    & \textbf{70.1 $\pm$ 1.8} & \textbf{69.6 $\pm$ 2.1} & \textbf{69.5 $\pm$ 1.9} & \textbf{75.2 $\pm$ 1.2} & \textbf{38.3 $\pm$ 3.6} & \textbf{5.24} \\
\midrule

\multirow{4}{*}{COVID-19}
 & LogReg & 72.1 $\pm$ 31.3 & 72.0 $\pm$ 31.3 & 71.9 $\pm$ 31.2 & 77.8 $\pm$ 33.9 & 58.3 $\pm$ 24.8 & \textbf{0.92} \\
 & SVM    & 72.1 $\pm$ 31.5 & 71.9 $\pm$ 31.4 & 71.9 $\pm$ 31.4 & 78.2 $\pm$ 34.1 & 58.2 $\pm$ 25.1 & 2.70 \\
 & MLP    & 75.0 $\pm$ 32.9 & 74.9 $\pm$ 32.8 & 74.9 $\pm$ 32.8 & 80.9 $\pm$ 35.6 & 64.2 $\pm$ 27.9 & 4.90 \\
 & GraphSAGE & \textbf{90.5 $\pm$ 0.8} & \textbf{90.5 $\pm$ 0.8} & \textbf{90.5 $\pm$ 0.8} & \textbf{96.6 $\pm$ 0.4} & \textbf{81.0 $\pm$ 1.6} & 7.18 \\
\midrule

\multirow{4}{*}{FakeNewsNet}
 & LogReg & 60.4 $\pm$ 26.3 & 59.6 $\pm$ 25.7 & 60.0 $\pm$ 25.9 & 56.5 $\pm$ 24.2 & 17.9 $\pm$ 7.1 & 6.71 \\
 & SVM    & 60.1 $\pm$ 25.6 & 64.5 $\pm$ 28.1 & 58.7 $\pm$ 25.5 & 55.6 $\pm$ 23.9 & 11.8 $\pm$ 4.8 & \textbf{6.23} \\
 & MLP    & 66.9 $\pm$ 29.3 & 68.3 $\pm$ 29.9 & 66.4 $\pm$ 28.8 & 65.7 $\pm$ 28.4 & 33.0 $\pm$ 13.7 & 9.93 \\
 & ChebNet & \textbf{79.3 $\pm$ 2.1} & \textbf{80.5 $\pm$ 1.8} & \textbf{79.1 $\pm$ 1.7} & \textbf{80.8 $\pm$ 1.0} & \textbf{42.9 $\pm$ 5.1} & 8.40 \\
\midrule

\multirow{4}{*}{Kaggle}
 & LogReg & 67.9 $\pm$ 29.8 & 67.9 $\pm$ 29.8 & 67.9 $\pm$ 29.8 & 74.3 $\pm$ 32.5 & 50.0 $\pm$ 21.7 & 16.79 \\
 & SVM    & 69.3 $\pm$ 30.3 & 69.2 $\pm$ 30.3 & 69.1 $\pm$ 30.3 & 75.7 $\pm$ 33.2 & 52.6 $\pm$ 22.8 & 205.07 \\
 & MLP    & 73.5 $\pm$ 32.1 & 73.3 $\pm$ 32.0 & 73.2 $\pm$ 32.0 & 78.7 $\pm$ 34.5 & 60.9 $\pm$ 26.3 & 14.21 \\
 & GraphSAGE & \textbf{96.8 $\pm$ 0.2} & 96.8 $\pm$ 0.2 & \textbf{96.8 $\pm$ 0.2} & \textbf{99.4 $\pm$ 0.0} & \textbf{93.5 $\pm$ 0.5} & \textbf{11.11} \\
\midrule

\multirow{4}{*}{LIAR}
 & LogReg & 49.4 $\pm$ 20.7 & 49.6 $\pm$ 20.9 & 49.4 $\pm$ 20.7 & 48.4 $\pm$ 19.6 & 11.5 $\pm$ 5.3 & \textbf{7.92} \\
 & SVM    & 45.9 $\pm$ 20.0 & 48.4 $\pm$ 21.2 & 40.0 $\pm$ 17.2 & 44.3 $\pm$ 19.1 & 2.7 $\pm$ 1.1 & 11.95 \\
 & MLP    & 48.3 $\pm$ 19.8 & 49.2 $\pm$ 20.4 & 48.1 $\pm$ 19.6 & 50.7 $\pm$ 20.5 & 9.3 $\pm$ 6.9 & 11.97 \\
 & GCN    & \textbf{60.7 $\pm$ 4.6} & \textbf{61.6 $\pm$ 3.6} & \textbf{59.2 $\pm$ 5.8} & \textbf{58.8 $\pm$ 4.8} & \textbf{18.9 $\pm$ 9.8} & 9.83 \\
\midrule

\multirow{4}{*}{MPID}
 & LogReg & 63.9 $\pm$ 28.1 & 63.4 $\pm$ 27.8 & 63.6 $\pm$ 27.9 & 67.9 $\pm$ 29.8 & 35.8 $\pm$ 15.5 & 13.27 \\
 & SVM    & 65.2 $\pm$ 28.0 & 65.4 $\pm$ 28.1 & 65.0 $\pm$ 27.8 & 69.7 $\pm$ 30.2 & 38.4 $\pm$ 15.3 & 321.47 \\
 & MLP    & 68.6 $\pm$ 30.0 & 68.6 $\pm$ 29.9 & 68.5 $\pm$ 30.0 & 75.3 $\pm$ 33.0 & 46.3 $\pm$ 19.9 & \textbf{12.23} \\
 & ChebNet & \textbf{88.9 $\pm$ 0.5} & \textbf{88.9 $\pm$ 0.5} & \textbf{88.6 $\pm$ 0.5} & \textbf{95.0 $\pm$ 0.6} & \textbf{74.0 $\pm$ 1.2} & 20.68 \\
\midrule

\multirow{4}{*}{WELFake}
 & LogReg & 63.4 $\pm$ 27.7 & 63.4 $\pm$ 27.7 & 63.4 $\pm$ 27.7 & 69.8 $\pm$ 30.5 & 41.0 $\pm$ 17.6 & \textbf{11.93} \\
 & SVM    & 63.5 $\pm$ 27.8 & 63.5 $\pm$ 27.8 & 63.5 $\pm$ 27.8 & 69.9 $\pm$ 30.6 & 41.3 $\pm$ 17.8 & 230.85 \\
 & MLP    & 67.0 $\pm$ 29.1 & 66.9 $\pm$ 29.1 & 66.8 $\pm$ 29.1 & 74.0 $\pm$ 32.4 & 48.1 $\pm$ 20.4 & 15.59 \\
 & GraphSAGE & \textbf{91.9 $\pm$ 0.2} & \textbf{91.9 $\pm$ 0.2} & \textbf{91.9 $\pm$ 0.2} & \textbf{97.2 $\pm$ 0.0} & \textbf{83.8 $\pm$ 0.4} & 16.21 \\
\bottomrule
\end{tabular}}
\end{table*}

\subsection{RQ2: Which classic GNNs work best across datasets?}
Table~\ref{tab:best_gnn_per_dataset} summarizes the best-performing classic GNN for each dataset after hyperparameter optimization, selected based on F1 score. The results indicate that no single GNN architecture dominates across all datasets.

GraphSAGE performs best on larger and more homogeneous datasets such as Kaggle, WELFake, and COVID-19, likely due to its effective neighborhood aggregation and scalability. ChebNet achieves the strongest performance on FakeNewsNet and MPID, where local structural information appears particularly important. Attention-based models such as GAT are most effective on CLICK-ID, which consists of short texts with subtle semantic distinctions.

These findings suggest that architectural choices play a crucial role and that multiple classic GNNs can serve as strong baselines, depending on the dataset characteristics. Different classic GNN architectures excel on different datasets, and no single model is universally optimal across all settings.
\begin{table}[t]
\centering
\caption{Best-performing classic GNN per dataset, selected by F1 score after
hyperparameter optimization. Results are reported as mean $\pm$ standard deviation.}
\label{tab:best_gnn_per_dataset}
\small
\setlength{\tabcolsep}{5pt}
\begin{tabular}{l l r r r p{1.5cm}}
\toprule
\textbf{Dataset} & \textbf{Best GNN} &
\textbf{Precision (\%)} &
\textbf{Recall (\%)} &
\textbf{F1 (\%)} &
\textbf{Inference (ms)} \\
\midrule
Kaggle       & GraphSAGE  & 96.8 $\pm$ 0.2 & 96.8 $\pm$ 0.2 & 96.8 $\pm$ 0.2 & 11.11 \\
WELFake      & GraphSAGE  & 91.9 $\pm$ 0.2 & 91.9 $\pm$ 0.2 & 91.9 $\pm$ 0.2 & 16.21 \\
COVID-19     & GraphSAGE  & 90.5 $\pm$ 0.8 & 90.5 $\pm$ 0.8 & 90.5 $\pm$ 0.8 & 7.18 \\
MPID         & ChebNet    & 88.9 $\pm$ 0.5 & 88.9 $\pm$ 0.5 & 88.6 $\pm$ 0.5 & 20.68 \\
FakeNewsNet  & ChebNet    & 79.3 $\pm$ 2.1 & 80.5 $\pm$ 1.8 & 79.1 $\pm$ 1.7 & 8.40 \\
CLICK-ID     & GAT        & 70.1 $\pm$ 1.8 & 69.6 $\pm$ 2.1 & 69.5 $\pm$ 1.9 & 5.24 \\
LIAR         & GCN        & 60.7 $\pm$ 4.6 & 61.6 $\pm$ 3.6 & 59.2 $\pm$ 5.8 & 9.83 \\
\bottomrule
\end{tabular}
\end{table}

\subsection{RQ3: Do classic GNNs remain effective in low-resource settings?}
Table~\ref{tab:rq3_low_resource} evaluates the robustness of classic GNNs under reduced training data availability by reporting F1 scores when only 10\%, 20\%, and 30\% of the training data are used.

Across all datasets, classic GNNs retain a substantial portion of their full-data performance even in low-resource settings. The observed F1 drop between 30\% and 10\% training data remains below 7 percentage points for all datasets and is negligible for some, such as LIAR. This robustness suggests that relational inductive bias allows GNNs to leverage graph structure to compensate for limited labeled data. Classic GNNs remain effective in low-resource settings, exhibiting only modest performance degradation when trained on a small fraction of the available data.
\begin{table}[t]
\centering
\caption{Effect of training data size on the performance of classic GNNs (F1 score).}
\label{tab:rq3_low_resource}
\begin{tabular}{lc@{ }c@{ }cp{2.5cm}}
\toprule
Dataset & F1@10\% & F1@20\% & F1@30\% & Drop (\%) \\
\midrule
CLICK-ID     & 65.0 & 69.0 & 69.5 & 6.5 \\
COVID-19     & 87.2 & 90.3 & 90.5 & 3.6 \\
FakeNewsNet  & 74.2 & 79.1 & 79.1 & 6.2 \\
Kaggle       & 93.5 & 95.7 & 96.8 & 3.4 \\
LIAR         & 59.2 & 57.3 & 58.8 & 0.7 \\
MPID         & 85.6 & 88.6 & 88.4 & 3.2 \\
WELFake      & 87.5 & 89.9 & 91.9 & 4.8 \\
\bottomrule
\end{tabular}

\end{table}

\subsection{RQ4: What is the performance–efficiency trade-off?}
\begin{figure}[t]
    \centering
    \includegraphics[width=0.8\linewidth]{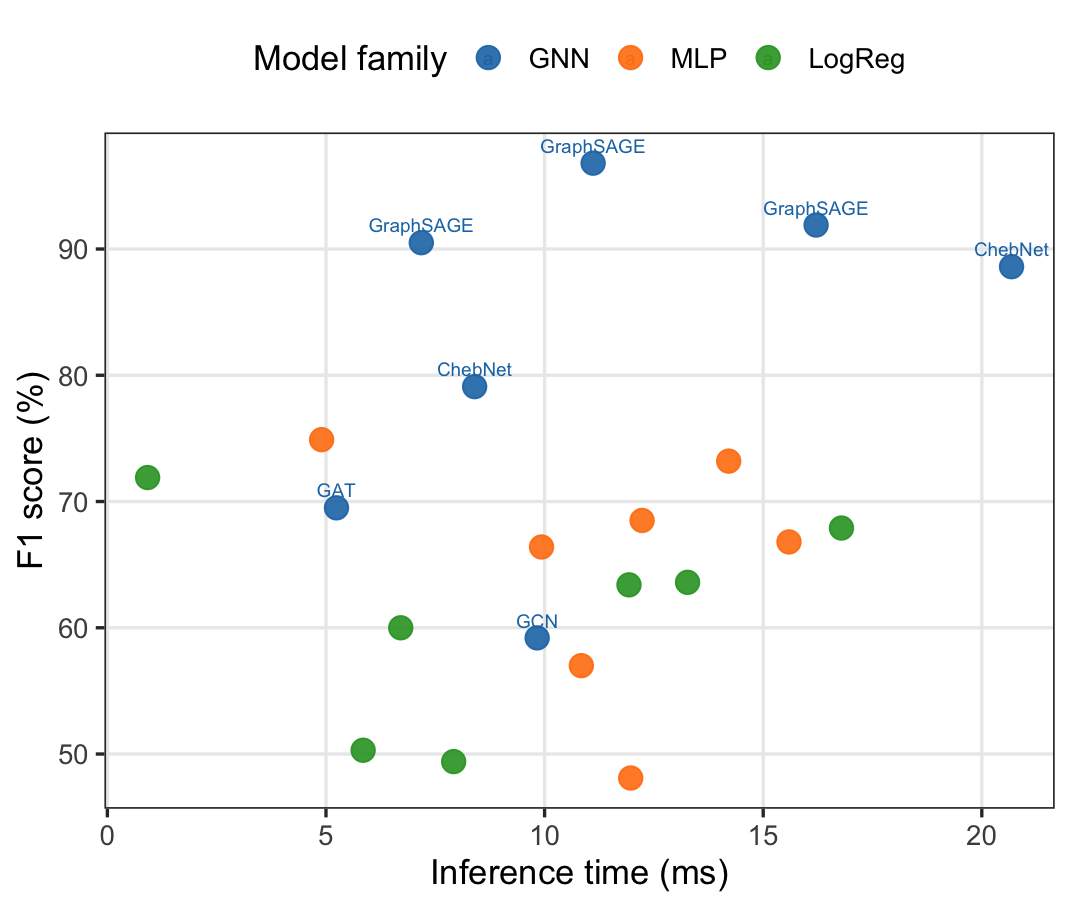}
    \caption{
    Performance--efficiency trade-off across learning algorithms.
    Each point corresponds to the best-performing configuration of a model on a given dataset.
    The x-axis reports inference time (ms), while the y-axis shows F1 score (\%).
    All GNN variants are shown using the same color to emphasize family-level behavior,
    with algorithm names annotated for clarity.
    }
    \label{fig:performance_efficiency_tradeoff}
\end{figure}

Figure~\ref{fig:performance_efficiency_tradeoff} visualizes the performance–efficiency trade-off between classic GNN architectures and representative non-graph baselines. Each point corresponds to the best-performing configuration of a model on a given dataset, with the x-axis indicating inference latency and the y-axis denoting F1 score. Models are grouped by family, with all GNN variants shown in the same color to emphasize family-level behavior while retaining algorithm-level labels.

Several clear trends emerge from the figure. Logistic Regression consistently achieves the lowest inference latency, reflecting its fixed-cost linear inference, but yields substantially lower detection performance across all datasets. MLP-based models provide moderate performance improvements over linear baselines; however, they remain consistently inferior to graph-based approaches, even when operating at comparable or higher inference costs.

In contrast, classic GNNs achieve pronounced gains in detection performance, frequently improving F1 scores by more than 20 percentage points relative to MLPs, while maintaining inference times within a practical range of a few milliseconds. Importantly, GNN inference costs remain comparable to those of MLPs, indicating that the observed performance improvements are not obtained at the expense of prohibitive computational overhead.

We exclude SVM-based classifiers from this analysis, as their inference complexity depends on the number of support vectors, which varies significantly across datasets and hyperparameter settings. This property complicates fair latency comparisons with models exhibiting fixed-cost inference, such as Logistic Regression, MLPs, and GNNs, particularly in large-scale settings.

\section{Discussion}
\label{sec:discussion}

This work revisits the effectiveness of classic graph neural networks for misinformation detection under controlled and practically oriented experimental conditions. Rather than introducing new modeling architectures, the study aims to clarify the circumstances under which lightweight GNNs remain competitive with strong non-graph baselines and to assess how their performance–efficiency characteristics compare with more complex modeling approaches.

Across all evaluated datasets, the results demonstrate that classic GNNs consistently benefit from relational inductive bias, even when applied to straightforward TF--IDF representations and shallow graph constructions. In comparison to Logistic Regression, SVM, and MLP baselines, GNNs achieve markedly higher F1 and MCC scores, indicating not only improved predictive accuracy but also more reliable performance under class imbalance. Notably, these improvements are observed across datasets that vary substantially in size, language, and domain, suggesting that the gains are robust rather than dataset-specific.

Equally important, the observed performance improvements do not come at the cost of excessive computational overhead. Inference times for GNNs remain comparable to those of MLPs and are frequently lower than those of SVMs on larger datasets. It suggests that the primary advantage of classic GNNs in this setting is their ability to efficiently exploit relational structure, rather than their increased model complexity. From a practical perspective, this makes lightweight GNNs well-suited for deployment in scenarios where latency, scalability, or limited computational resources are key constraints.

The dataset-level analysis further indicates that no single GNN architecture consistently dominates across all benchmarks. GraphSAGE performs particularly well on larger and more homogeneous datasets. At the same time, ChebNet and GAT are more effective in settings where local graph structure or fine-grained semantic distinctions play a critical role. These findings suggest that classic GNNs should be viewed as a set of complementary modeling choices rather than as a single uniform solution.

\paragraph{On Comparisons with Transformer-based Models.}
A natural question is whether direct comparisons with large Transformer-based models are necessary or appropriate in this context. Although Transformer architectures have demonstrated strong performance in misinformation detection~ \cite{transformers10967607,transformersalarfaj2026real,transformersvenkataramanan2026hca}, they rely on fundamentally different assumptions, including deep contextual language modeling, extensive pre-training, and substantial computational resources. Prior work has shown that these characteristics limit their applicability in low-resource, multilingual, or latency-sensitive settings.

In contrast, the focus of this study is on scenarios where simplicity, interpretability, and computational efficiency are central considerations. Directly comparing classic GNNs with large Transformer models would confound architectural complexity with representational capacity, obscuring the specific contribution of graph-based relational modeling. Accordingly, Transformer-based approaches are treated as a reference point rather than as primary baselines, and the evaluation emphasizes controlled comparisons against non-graph models operating on identical feature representations. This design choice enables the assessment of the contribution of relational inductive bias in isolation.

Importantly, these findings do not suggest that classic GNNs supersede \\Transformer-based approaches in all settings. Instead, they indicate that for many practical misinformation detection tasks—particularly those involving limited computational resources or multilingual data—classic GNNs offer a balanced and effective alternative between simple linear models and highly complex neural architectures.

\section{Conclusion}
\label{sec:conclusion}

This paper presented a systematic, large-scale evaluation of classic graph neural networks for misinformation detection across seven datasets spanning multiple languages, domains, and label distributions. By benchmarking lightweight GNN architectures against strong non-graph baselines under identical feature representations, we demonstrated that classic GNNs remain both effective and efficient in contemporary misinformation detection settings.

Our results show that classic GNNs consistently outperform Logistic Regression, SVM, and MLP baselines across all datasets, achieving higher F1 scores and MCC values while maintaining practical inference times. These gains persist in low-resource settings, where GNNs exhibit only modest performance degradation even when trained on as little as 10\% of the available data. This robustness highlights the value of relational inductive bias when labeled data is scarce.

Crucially, the study demonstrates that meaningful performance improvements can be achieved without the need for increasingly complex architectures or heavy reliance on large language models. Carefully designed graph constructions combined with classic GNNs already provide strong baselines that are competitive, interpretable, and computationally efficient. This finding challenges the prevailing trend toward architectural complexity and underscores the importance of reassessing simpler methods under fair and controlled conditions. Future work may explore hybrid settings in which classic GNNs are combined with lightweight contextual embeddings, investigate alternative graph construction strategies beyond k-NN similarity, or extend the evaluation to streaming and early-detection scenarios. More broadly, we hope this work encourages the community to treat classic GNNs not as outdated baselines, but as viable and often preferable solutions for real-world misinformation detection systems. Future work includes exploring another architecture optimization framework based on Genetic Programming~\cite{10.1145/3712255.3734278,10.1145/3712255.3734538}, Evolutionary Strategies~\cite{10.1145/3746252.3761661}, and bio-inspired methods~\cite{badap-agh:163984}. This should enable more flexible design and optimization~\cite{10.1145/3746252.3761661,11214657}. The promising experimental results may be valuable for researchers working on Neural Architecture Search.

\section*{Acknowledgments}
The authors gratefully acknowledge the Polish high-performance computing infrastructure PLGrid (HPC Center: ACK Cyfronet AGH) for providing computer facilities and support within computational grant no. PLG/2025/018082 (M.K.).

A.W. and S.K. were funded by the European Union under the Horizon Europe project OMINO (grant agreement no. 101086321). Views and opinions expressed are, however, those of the authors only and do not necessarily reflect those of the European Union or the European Research Executive Agency. Neither the European Union nor the European Research Executive Agency can be held responsible for them.
Additionally, A.W. and S.K. were co-financed by the Polish Ministry of Education and Science under the program International Co-Financed Projects.
%
%
 \bibliographystyle{splncs04}
 \bibliography{mybib}

\end{document}